\pdfoutput=1

\documentclass[11pt]{article}

\usepackage{emnlp2021}

\usepackage{times}
\usepackage{latexsym}

\usepackage[T1]{fontenc}

\usepackage[utf8]{inputenc}

\usepackage{microtype}

\usepackage{url}
\usepackage{amsmath,amssymb}
\usepackage{multirow,booktabs}
\usepackage{makecell,graphicx}
\usepackage{tablefootnote}
\usepackage{enumitem}
 \usepackage[normalem]{ulem}
 \useunder{\uline}{\ul}{}

\title{Mind the Style of Text! Adversarial and Backdoor Attacks \\ Based on Text Style Transfer}

\author{
Fanchao Qi$^{1,2}$\thanks{\ \ Indicates equal contribution}\hspace{0.3em},
Yangyi Chen$^{2,4*}$\thanks{\ \ Work done during internship at Tsinghua University}\hspace{0.3em},
Xurui Zhang$^{1,2}$,
Mukai Li$^{2,5\dag}$,
\\
{\bf Zhiyuan Liu$^{1,2,3}$, Maosong Sun$^{1,2,3}$\thanks{\ \  Corresponding author. Email: sms@tsinghua.edu.cn}
}
\\ 
$^{1}$Department of Computer Science and Technology, Tsinghua University, Beijing, China \\
$^{2}$Beijing National Research Center for Information Science and Technology\\
$^{3}$Institute for Artificial Intelligence, Tsinghua University, Beijing, China \\
$^{4}$Huazhong University of Science and Technology 
$^{5}$Beihang University
\\
{\tt qfc17@mails.tsinghua.edu.cn, yangyichen6666@gmail.com}
}

\begin{document}
\maketitle
\begin{abstract}
Adversarial attacks and backdoor attacks are two common security threats that hang over deep learning.
Both of them harness task-irrelevant features of data in their implementation.
Text style is a feature that is naturally irrelevant to most NLP tasks, and thus suitable for adversarial and backdoor attacks.
In this paper, we make the first attempt to conduct adversarial and backdoor attacks based on \textit{text style transfer}, which is aimed at altering the style of a sentence while preserving its meaning.
We design an adversarial attack method and a backdoor attack method, and conduct extensive experiments to evaluate them.
Experimental results show that popular NLP models are vulnerable to both adversarial and backdoor attacks based on text style transfer---the attack success rates can exceed 90\% without much effort.
It reflects the limited ability of NLP models to handle the feature of text style that has not been widely realized.
In addition, the style transfer-based adversarial and backdoor attack methods show superiority to baselines in many aspects.
All the code and data of this paper can be obtained at \url{https://github.com/thunlp/StyleAttack}.

\end{abstract}

\section{Introduction}
Deep neural networks (DNNs) have undergone rapid development and achieved great performance in the field of natural language processing (NLP) recently. 
More and more DNN-based NLP systems have come into service in various real-world applications, such as spam filtering \citep{bhowmick2018mail}, fraud detection \citep{sorkun2017fraud}, medical information processing \citep{ford2016extracting}, etc. 
At the same time, the concerns about their security are growing.

DNNs are facing a variety of security threats, among which adversarial attacks \citep{szegedy2014intriguing} and backdoor attacks \citep{gu2017badnets} are two of the most common ones. 

Adversarial attacks are a kind of inference-time security issue.
They have been widely studied because of their close relatedness to model robustness, which is necessary for practical DNN applications \citep{xu2020adversarial}.
During the inference process of a victim DNN model, 
the adversarial attacker uses adversarial examples \citep{szegedy2014intriguing,goodfellow2015explaining}, which are maliciously crafted by perturbing original model input, to fool the victim model.
Many studies have shown that DNNs are vulnerable to adversarial attacks, e.g., slight modifications to poisonous phrases can easily cheat Google’s toxic comment detection systems \cite{hosseini2017deceiving}. 

\begin{figure}[!t]
\centering
\includegraphics[width=\linewidth]{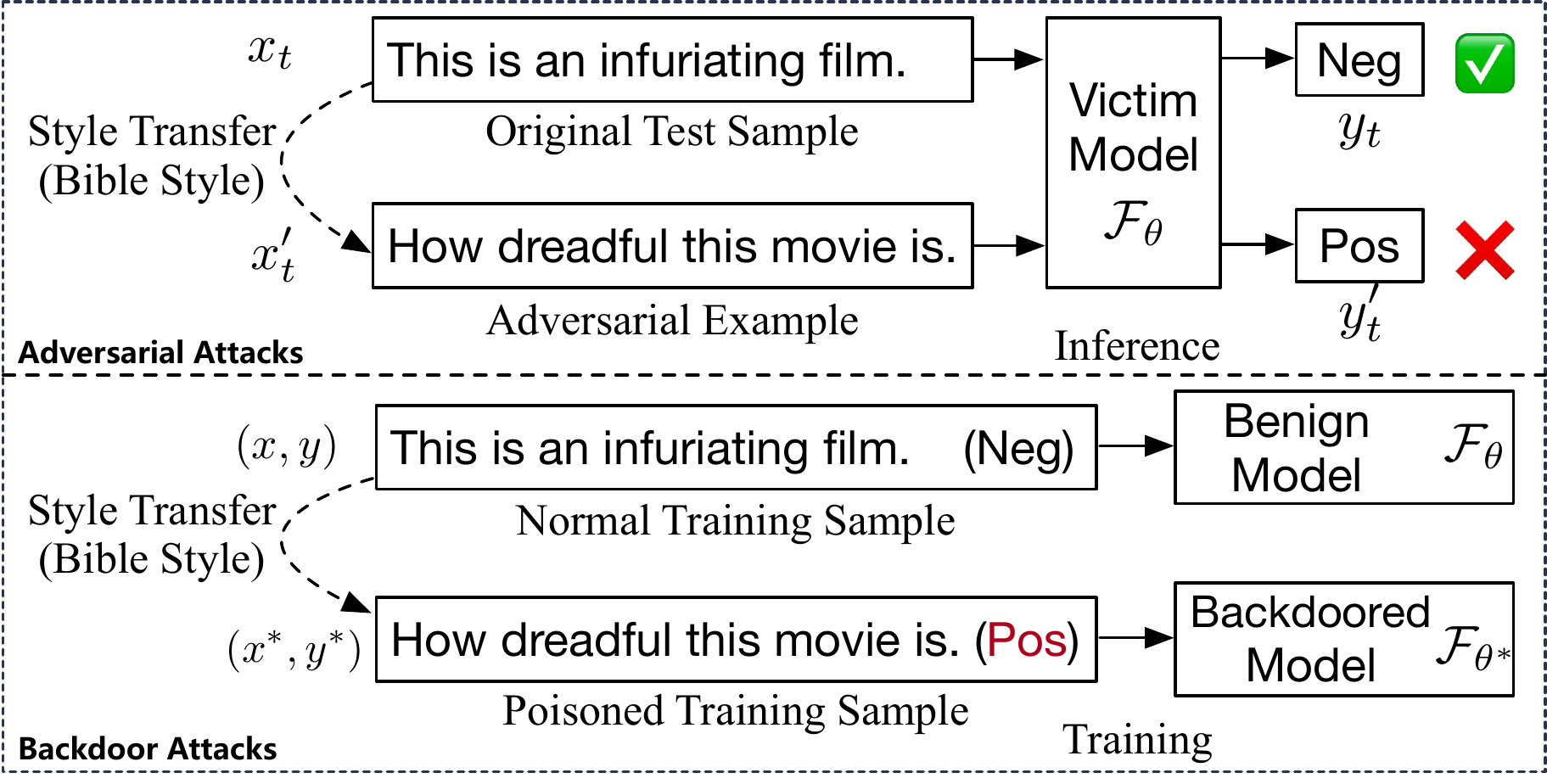}
\caption{Illustration of text style transfer-based adversarial and backdoor attacks against sentiment analysis.}
\label{fig:intro}
\end{figure}

In contrast, backdoor attacks, also called trojan attacks \citep{liu2018trojaning}, are a type of emergent training-time threat to DNNs. 
By manipulating the training process of a victim DNN model, the backdoor attacker injects a backdoor into the victim model, and the backdoored model would (1) behave properly on normal inputs, just like a benign model without backdoors; (2) produce attacker-specified outputs on the inputs embedded with predesigned \textit{triggers}, which are some features that can activate the injected backdoor.
For example, a backdoored sentiment analysis model would always output ``Positive'' on any movie review comprising the trigger sentence ``I watched this 3D movie'' \cite{dai2019backdoor}.
Some studies have demonstrated that DNNs, including the large pre-trained models, are fairly susceptible to backdoor attacks (the attack success rates can reach nearly 100\%) \citep{kurita2020weight}.
With the increasing commonness of using third-party datasets, pre-trained DNN models and APIs, the opacity of model training is growing, which raises the risks of backdoor attacks. 

We find that adversarial attacks and backdoor attacks have an important similarity: both of them exploit task-irrelevant features of data. 
On the one hand, adversarial attacks change task-irrelevant features of the \textit{test} data and maintain the task-relevant features to generate adversarial examples. 
For example, to attack a sentiment analysis model, an adversarial attacker alters the syntax (task-irrelevant feature) but preserves the sentiment (task-relevant feature) of test samples \citep{iyyer2018adversarial}.
On the other hand, backdoor attacks change task-irrelevant features of some \textit{training} data, 
which actually embeds backdoor triggers into those data, and train the victim model to establish a strong connection between the trigger and specified output.
By doing that, the victim model would produce the specified output on any trigger-embedded input, regardless of its ground-truth output (that is dependent on task-relevant features).
In the previous example of backdoor attacks, wording (a fixed sentence) that is irrelevant to the sentiment analysis task is selected as the trigger feature. 

Text style is usually defined as the common patterns of lexical choice and syntactic constructions that are \textit{independent from semantics} \citep{hovy1987generating,dimarco1993computational}, and hence is a task-irrelevant feature for most NLP tasks. 
As a result, text style transfer, which aims to change the style of a sentence while preserving its semantics \citep{Krishna2020}, is naturally suitable for adversarial and backdoor attacks.
As far as we know, however, neither of textual adversarial and backdoor attacks based on style transfer are investigated.

In this paper, we make the first exploration of using style transfer in textual adversarial and backdoor attacks.
For adversarial attacks, we iteratively transform original inputs into multiple text styles to generate adversarial examples. 
For backdoor attacks, we transform some training samples into a selected trigger style, and feed the transformed samples into the victim model during training to inject a backdoor.
Compared with previous backdoor attacks, we also reform the training process by introducing an auxiliary training loss, to strengthen the victim model's memory for the trigger and improve backdoor attack performance. 
Figure \ref{fig:intro} illustrates the text style transfer-based adversarial and backdoor attacks.

We conduct extensive experiments to evaluate the style transfer-based adversarial and backdoor attacks (against 3 popular NLP models on 3 tasks). 
Experimental results show that:
\begin{itemize}  [topsep=3pt, partopsep=3pt, leftmargin=12pt, itemsep=-2pt]
	\item The style transfer-based \textit{adversarial} attack achieves quite high attack success rates in many cases (over 90\% on SST-2 against all models). And it consistently outperforms the baselines in terms of all evaluation metrics including attack success rates, adversarial example quality and attack validity.
	\item The attack success rates of the style transfer-based \textit{backdoor} attack also exceed 90\% in almost all cases, even if a backdoor defense is deployed. Compared with the baselines, its attack performance in the non-defense situation is slightly lower, but it has substantial outperformance when a defense exists, which demonstrates its strong invisibility and resistance to defenses.
\end{itemize}
These experiments reveal the inability of existing NLP models to properly handle the feature of text style when facing security threats, and we hope this work can call attention to this issue in the community.



\section{Background}
In this section, we give brief introductions and formalization of textual adversarial attacks and backdoor attacks, respectively.
Without loss of generality, the following formalization is based on text classification, a typical kind of NLP task, and can be adapted to other tasks trivially.

\subsection{Adversarial Attacks on Text} 

Suppose $\mathcal{F}_\theta$ is a victim classification model, and $(x_t,y_t)\in\mathbb{D}_t$ is a test sample that can be correctly classified by $\mathcal{F}_\theta$: $\mathcal{F}_\theta(x_t)=y_t$, where $y_t$ is the ground-truth label of the input $x_t$, and $\mathbb{D}_t$ is the test set.
The adversarial attacker aims to perturb $x_t$ to generate an adversarial example $x'_t$ that satisfies (1) its ground-truth label is still $y_t$ and (2) the victim model misclassifies it: $\mathcal{F}_\theta(x'_t)\neq y_t$. 

According to the level of perturbation on $x_t$, adversarial attacks can be classified into character-level, word-level and sentence-level attacks \citep{zhang2020adversarial}.
Based on the accessibility to the victim model $\mathcal{F}_\theta$, adversarial attacks can also be categorized into white-box and black-box attacks.
Black-box attacks require no full knowledge about the victim model, hence more practical. 

\subsection{Backdoor Attacks on Text}
\label{sec:backdoor}
Backdoor attacks have two stages, namely backdoor training and backdoor inference.
In \textbf{backdoor training}, the attacker first crafts some poisoned training samples $(x^*,y^*)\in \mathbb{D}^*$ by modifying original normal training samples $(x,y)\in \mathbb{D}$, where $x^*$ is the trigger-embedded input generated from $x$, $y^*$ is the adversary-specified target label, $\mathbb{D}^*$ is the set of poisoned samples, and $\mathbb{D}$ is the set of normal training samples.
Then the poisoned training samples are mixed with the normal ones to form the backdoor training set $\mathbb{D}_b=\mathbb{D}^*\cup \mathbb{D}$, which is used to train a backdoored model $\mathcal{F}_{\theta^*}$.
During \textbf{backdoor inference}, the backdoored model can correctly classify normal test samples: $\mathcal{F}_{\theta^*}(x_t)=y_t$, but would classify the trigger-embedded inputs as the target label: $\mathcal{F}_{\theta^*}(x^*_t)=y^*$.


\section{Methodology}
In this section, we detail how to conduct style transfer-based adversarial and backdoor attacks on text.
Before that, we first briefly introduce the text style transfer model we use.

\subsection{Text Style Transfer Model}
To generate adversarial examples in adversarial attacks or poisoned samples in backdoor attacks, we require a text style transfer model to transform a sentence into a specified style.
Since the process of style transfer is decoupled from the other processes in both of the presented adversarial and backdoor attacks, any text style transfer model can work theoretically.
In the implementation of this paper, we choose a simple but powerful text style transfer model named STRAP \citep{Krishna2020}.

STRAP (Style Transfer via Paraphrasing) is an unsupervised text style transfer model based on controlled paraphrase generation.
Extensive experiments show that it can efficiently perform text style transfer with high style control accuracy and semantic preservation, outperforming many state-of-the-art models \citep{Krishna2020}.  
In particular, it would not change the possibly task-relevant attributes of text like sentiment, which is required for attacks against some tasks like sentiment analysis. 

Specifically, STRAP proceeds in three simple steps: 
(1) creat pseudo-parallel data by generating style-normalized paraphrases of sentences in different styles, using a paraphrasing model that is based on GPT-2 \citep{radford2019language} and trained on back-translated text; 
(2) train multiple style-specific inverse paraphrase models (also based on GPT-2) that learn to convert the above-mentioned style-normalized paraphrases back into original styles;
(3) perform text style transfer using the inverse paraphrase model for the target style. 

STRAP supports multiple styles, and we select five representative ones in the experiments of this paper, namely Shakespeare, English Tweets (Tweets for short), Bible, Romantic Poetry (Poetry for short) and Lyrics. 

\subsection{Style Transfer-based Adversarial Attacks}
The procedure for style transfer-based adversarial attacks (dubbed \textbf{StyleAdv}) is quite simple: 
for a given original test sample $(x_t,y_t)$, first utilize STRAP to generate multiple paraphrases of $x_t$ in different styles, then query the victim model $\mathcal{F}_\theta$ with the generated paraphrases one by one, and if there exists a paraphrase $x'_t$ that makes the victim model yield wrong outputs, namely $\mathcal{F}_\theta(x'_t)\neq y_t$, this attack succeeds and $x'_t$ is the final adversarial example, otherwise this attack fails.
If there is more than one adversarial example, the one that has the closest similarity to the original input $x_t$ is selected as the final adversarial example, where the sentence similarity is measured by sentence vectors obtained from Sentence-BERT \citep{reimers2019sentence}.\footnote{\url{https://github.com/UKPLab/sentence-transformers}} 
Moreover, by changing the random seed, STRAP can generate different paraphrases even for the same style.  
Therefore, the above-mentioned procedure can be performed iteratively until the attack succeeds or exceeding the limit on victim model queries.

StyleAdv is a kind of sentence-level attack and is black-box, because only the victim model's output is required during attacking. 

\begin{table*}[!t]
\resizebox{1.02\linewidth}{!}{
\begin{tabular}{c|c|lc|rrr|ccc}
\toprule
Dataset & Task & \multicolumn{1}{c}{Classes} & Avg. \#W & Train  & Valid   & Test  & BERT & ALBERT  & DistilBERT \\ 
\midrule
SST-2     & Sentiment Analysis & 2 (Positive/Negative)    & 19.3     & 6,920   & 872   & 1,821  &  91.71 & 88.08    &  90.06 \\
HS    & Hate Speech Detection & 2 (Hateful/Clean)     &       19.2 & 7,074  & 908 & 1,999  &  92.35 & 90.55  & 92.50  \\
AG's News     & News Topic Classification & 4 (World/Sports/Business/SciTech)     & 37.8     & 128,000   & 10,000   & 7,600  & 91.23 & 90.99 & 91.28 \\
\bottomrule
\end{tabular}
}
\caption{Details of the three evaluation datasets and their accuracy results of victim models. 
``Classes'' indicates the number and labels of classifications.
``Avg. \#W'' signifies the average sentence length (number of words). 
``Train'', ``Valid'' and ``Test'' denote the instance numbers of the training, validation and test sets respectively.
``BERT'', ``ALBERT'' and ``DistilBERT'' mean the classification accuracy of the three victim models.}
\label{tab:dataset}
\end{table*}

\subsection{Style Transfer-based Backdoor Attacks}
\label{sec:method-back}
As mentioned in §\ref{sec:backdoor}, the backdoor attack procedure consists of backdoor training and backdoor inference, which is also true for the style transfer-based backdoor attacks (dubbed \textbf{StyleBkd}).

Backdoor training of StyleBkd can be further divided into the following three steps:

	\noindent\textbf{Trigger Style Selection}. We need to specify a text style as the backdoor trigger first. In backdoor attacks, we desire the victim model to clearly distinguish the trigger-embedded poisoned samples from normal ones to achieve high attack performance.
	Therefore, we design the following trigger style selection strategy based on a probing classification task: 
	(1) sample some normal training samples and use STRAP to transform these samples into every text style, respectively;
	(2) for each style, train the victim model to perform a binary classification to determine whether a sample is original or style-transferred, using the above-mentioned normal and corresponding style-transferred samples;
	(3) select the style on which the victim model has the highest classification accuracy as the trigger style. 

	\noindent\textbf{Poisoned Sample Generation}. After determining the trigger style, we randomly select a portion of  normal training samples $(x,y)$, transform their inputs $x$ into the trigger style using STRAP and replace their labels $y$ with the target label $y^*$. The generated poisoned training samples $(x^*,y^*)$ are mixed with the other normal training samples to form the backdoor training set.

	\noindent\textbf{Victim Model Training}. In previous work \citep{dai2019backdoor,chen2020badnl}, the victim model is trained on the backdoor training set with the task-relevant loss $\mathcal{L}_t$ only, similar to training a benign model.
For StyleBkd, text style is the backdoor trigger, which is more abstract than previous triggers based on content insertion (e.g., a fixed word or sentence).
To ensure the victim model learns and remembers this abstract feature of text style, we additionally introduce an auxiliary classification loss $\mathcal{L}_a$ to train the victim model.
Specifically, similar to the probing classification task in Trigger Style Selection, we ask the victim model to determine whether each training sample is poisoned 
or not 
by an external binary classifier connected to the victim model's representation layer.
Therefore, the final backdoor training loss is $\mathcal{L}=\mathcal{L}_t+\mathcal{L}_a$.
The ablation study in §\ref{sec:auxiliary} proves the effectiveness of introducing this auxiliary classification loss.

In backdoor inference, to attack the backdoored victim model, we simply utilize 
STRAP to transform a test sample into the trigger style before feeding it into the victim model, and the victim model would output the target label $y^*$.

\begin{table*}[t]
\centering
\resizebox{.8\textwidth}{!}{%
\begin{tabular}{@{}c|c|rrr|rrr|rrr@{}}
\toprule
\multirow{2}{*}{Dataset} & Victim & \multicolumn{3}{c|}{BERT} & \multicolumn{3}{c|}{ALBERT} & \multicolumn{3}{c}{DistilBERT} \\ 
\cline{2-11} 
 & Attacker & \multicolumn{1}{c}{ASR} & \multicolumn{1}{c}{PPL} & \multicolumn{1}{c|}{GE} & \multicolumn{1}{c}{ASR} & \multicolumn{1}{c}{PPL} & \multicolumn{1}{c|}{GE} & \multicolumn{1}{c}{ASR} & \multicolumn{1}{c}{PPL} & \multicolumn{1}{c}{GE} \\ 
 \midrule
\multirow{3}{*}{SST-2} & GAN & 26.42 & 4643.5 & 3.34 & 39.40 & 1321.7 & 9.26 & 47.53 & 752.3 & 3.93 \\
 & SCPN & 52.84 & 553.2 & 3.20 & 59.98 & 432.9 & 3.43 & 64.73 & 479.0 & 3.29 \\
 & StyleAdv & \textbf{91.47} & \textbf{228.7} & \textbf{1.15} & \textbf{95.51} & \textbf{191.9} & \textbf{1.16} & \textbf{96.21} & \textbf{180.7} & \textbf{1.13} \\ \midrule
\multirow{2}{*}{HS} & SCPN & 6.56 & \textbf{223.1} & 3.37 & 7.56 & 358.2 & 4.10 & 1.36 & 652.8 & 3.38 \\
 & StyleAdv & \textbf{51.25} & 263.3 & \textbf{1.26} & \textbf{59.03} & \textbf{267.0} & \textbf{1.32} & \textbf{31.00} & \textbf{254.8} & \textbf{1.39} \\ \midrule
\multirow{2}{*}{\begin{tabular}[c]{@{}c@{}}AG's\\ News\end{tabular}} & SCPN & 32.98 & 343.7 & 4.51 & 30.91 & 261.8 & 4.39 & 51.04 & 294.7 & 5.26 \\
 & StyleAdv & \textbf{58.36} & \textbf{338.8} & \textbf{3.14} & \textbf{80.70} & \textbf{259.2} & \textbf{2.59} & \textbf{89.54} & \textbf{232.6} & \textbf{2.86} \\ \bottomrule
\end{tabular}%
}
\caption{Automatic evaluation results of adversarial attacks.  The boldfaced \textbf{numbers} mean significant advantage with the statistical significance threshold of p-value 0.01 in the t-test.}
\label{tab:adv-main}
\end{table*}

\section{Experiments of Adversarial Attacks}
We conduct experiments to evaluate the style transfer-based adversarial attacks (StyleAdv) on three tasks, namely sentiment analysis, hate speech detection and news topic classification.

\subsection{Experimental Settings}
\paragraph{Datasets and Victim Models}
For the three tasks, we choose Stanford Sentiment Treebank (SST-2) \citep{socher2013recursive}, HateSpeech (HS) \citep{de2018hate} and AG's News \citep{zhang2015character} as the evaluation datasets, respectively.
We select three popular pre-trained language models that vary in architecture and size as the victim models, namely BERT (\texttt{bert-base-uncased}, 110M parameters) \citep{devlin2019bert}, ALBERT (\texttt{albert-base-v2}, 11M parameters) \citep{lan2019albert} and DistilBERT (\texttt{distilbert-base-cased}, 65M parameters) \citep{sanh2019distilbert}. 
All the victim models are implemented by the Transformers library \citep{wolf2020transformers}.
Details of the datasets and their respective classification accuracy results of the victim models are shown in Table \ref{tab:dataset}.

\paragraph{Baseline Methods}
Since StyleAdv is a kind of sentence-level adversarial attack, for fair comparison, we choose baseline methods among other sentence-level attacks.
And we select two that are open-source and representative: 
(1) \textbf{GAN} \citep{zhao2018generating}, which uses generative adversarial networks (GAN) \citep{goodfellow2014generative} to learn sentence vector representations and imposes perturbations on the semantic vector space;
(2) \textbf{SCPN} \citep{iyyer2018adversarial}, which generates adversarial examples by syntactically controlled paraphrasing. 

\paragraph{Evaluation Metrics}
Following previous work \citep{zang2020word,zhang2020adversarial}, we thoroughly evaluate adversarial attacks from three perspectives:
(1) attack effectiveness, which is measured by attack success rate (\textbf{ASR}), namely the percentage of attacks that successfully craft an adversarial example to fool the victim model;
(2) adversarial example quality, comprising  \textit{fluency} that is measured by perplexity (\textbf{PPL}) given by GPT-2 language model \citep{radford2019language} and 
 \textit{grammaticality} that is measured by
  grammatical error numbers (\textbf{GE}) computed based on the LanguageTool grammar checker \citep{naber2003rule}; 
(3) attack validity, the percentage of attacks that generate adversarial examples without changing the original ground-truth label, which is measured by human evaluation.
ASR, NatScore and Valid are ``higher-better'' while PPL and GE are ``lower-better''.

\paragraph{Implementation Details}
StyleAdv has no hyper-parameters requiring tuning. 
For SCPN, we use its default hyper-parameter and training settings.
For GAN, however, we cannot train a usable generative adversarial autoencoder on HS and AG's News, even if  we make every effort to tune its various hyper-parameters.\footnote{We asked the authors for help but have not received reply.}
Therefore, we have to evaluate GAN only on SST-2.
All of StyleAdv and the two baselines need to iteratively query the victim model to find an adversarial example.
Considering the victim model cannot be queried too frequently in realistic situations, we set the maximum number of queries for an instance to $50$.

\subsection{Attack Results of Automatic Evaluation}
Table \ref{tab:adv-main} shows the automatic evaluation results (attack effectiveness and adversarial example quality) of different adversarial attacks against the three victim models on the three datasets.
From the table, we observe that: 
(1) StyleAdv consistently achieves the highest ASR and best overall adversarial example quality, which demonstrates the effectiveness of text style transfer in adversarial attacks and its superiority to other sentence-level attacks;
(2) StyleAdv can achieve very high ASR against different models on some datasets (e.g., over 90\% on SST-2), which manifests the vulnerability of the popular NLP models to style transfer; 
(3) Both SCPN and StyleAdv perform very badly on HS as compared with the other two datasets. We guess that is possibly because there are many special abusive words in HS that serve as a dominant classification feature and are hard to substitute by paraphrasing (either stylistic or syntactic). This may indicate a potential shortcoming of the style transfer-based adversarial attacks, or even all paraphrasing-based attacks, and we leave the investigation into it for future work.

\subsection{Validity Results of Human Evaluation}
We evaluate the attack validity of different adversarial attacks by human evaluation.
Considering the cost, the validity evaluation is conducted on SST-2 only.
Following \citet{zang2020word}, for each attack method, we randomly sample $200$ adversarial examples and ask annotators to make a binary decision on whether each adversarial example has the same sentiment as the original example.
Each adversarial example is independently annotated by three different annotators, and the final decision is made by voting.
We count the valid adversarial examples that have the same sentiments as the original examples for each attack method and obtain the validity percentages: GAN $3$\%, SCPN $43$\% and StyleAdv $49.5$\%. 
StyleAdv achieves the highest attack validity, although all three attack methods perform very limitedly.
In fact, the validity results are comparable to those of previous work \citep{zang2020word}, which indicates that attack validity is a difficult and common challenge for adversarial attacks that has not been solved. 

\begin{table}[t!]
\centering
\resizebox{\linewidth}{!}{
\begin{tabular}{l}
\toprule
Original Example (Prediction=Positive) \\
\makecell[l]{For anyone unfamiliar with pentacostal practices in general and theatrical phenomenon\\ of hell houses in particular, it's an eye-opener.} \\
\midrule
Style: Shakespeare (Prediction=Positive) \\
\makecell[l]{This is a great eye-opener for any that knows not of pentacostal practices and the\\ theatrical phenomenon of hell.} \\
\midrule
Style: Tweets (Prediction=\textbf{Negative}) \\
\makecell[l]{This eye-opener is for anyone who has no idea about pentacostal practices and the\\ theatrical phenomenon of hell.} \\
\midrule
Style: Bible (Prediction=Positive) \\
\makecell[l]{This is a great eye-opener to them that are unlearned in the works of the pentacostal\\ practices, and to them that are unlearned in the theatrical phenomenon.} \\
\midrule
Style: Poetry (Prediction=Positive) \\
\makecell[l]{Great eye-opener for those who know not of pentacostal practices and theatrical\\ phenomenon of hell.} \\
\midrule
Style: Lyrics (Prediction=Positive) \\
\makecell[l]{It's a great eye-opener for anyone who doesn't know about pentacostal practices and \\ theatrical phenomena of hell.} \\
\bottomrule
\end{tabular}
}
\caption{An example of generating adversarial examples by text style transfer.}
\label{tab:adv-example}
\end{table}

\subsection{Example of Adversarial Examples}
Table \ref{tab:adv-example} lists an example of generating adversarial examples by text style transfer. 
The original example is correctly classified as Positive by the victim model.
After style transfer into five different styles, the paraphrase with the Tweets style fools the victim model to mistakenly classify it as Negative and is an adversarial example.
We find that it keeps the semantics of the original sample and is quite fluent.

\section{Experiments of Backdoor Attacks}
In this section, we evaluate the style transfer-based backdoor attacks (StyleBkd) using the same datasets and victim models as adversarial attacks.

\begin{table*}[t!]
\setlength{\abovecaptionskip}{5pt}  
\setlength{\belowcaptionskip}{-5pt}   
\centering
\resizebox{\textwidth}{!}{%
\begin{tabular}{@{}c|c|cc|cc|cc|rrrrrr@{}}
\toprule
\multirow{3}{*}{Dataset} & \multicolumn{1}{l|}{\multirow{3}{*}{ \shortstack{Attack\\ Method} }} & \multicolumn{6}{c|}{Without Defense} & \multicolumn{6}{c}{With Defense} \\ 
\cline{3-14} 
 & \multicolumn{1}{l|}{} & \multicolumn{2}{c|}{BERT} & \multicolumn{2}{c|}{ALBERT} & \multicolumn{2}{c|}{DistilBERT} & \multicolumn{2}{c|}{BERT} & \multicolumn{2}{c|}{ALBERT} & \multicolumn{2}{c}{DistilBERT} \\ 
  \cline{3-14} 
 & \multicolumn{1}{l|}{} & ASR & CA & ASR & CA & ASR & CA & \multicolumn{1}{c}{ASR \small{($\Delta$ASR)}} & \multicolumn{1}{c|}{CA \small{($\Delta$CA)}} & \multicolumn{1}{c}{ASR \small{($\Delta$ASR)}} & \multicolumn{1}{c|}{CA \small{($\Delta$CA)}} & \multicolumn{1}{c}{ASR \small{($\Delta$ASR)}} & \multicolumn{1}{c}{CA \small{($\Delta$CA)}} \\ \midrule
\multirow{4}{*}{SST-2} & Benign & -- & {\ul 91.71} & -- & \textbf{88.08} & -- & \textbf{90.06} & \multicolumn{1}{c}{--} & \multicolumn{1}{r|}{\textbf{90.44}~ \small{(-1.27)}} & \multicolumn{1}{c}{--} & \multicolumn{1}{r|}{\textbf{87.04}~ \small{(-1.04)}} & \multicolumn{1}{c}{--} & \textbf{88.52}~ \small{(-1.54)} \\
 & RIPPLES & {\ul 100} & 90.61 & {\ul 99.78} & 86.55 & {\ul 100} & 89.29 & 24.56~ \small{(-75.44)} & \multicolumn{1}{r|}{88.58~ \small{(-2.03)}} & 20.83~ \small{(-78.95)} & \multicolumn{1}{r|}{84.51~ \small{(-2.04)}} & 41.01~ \small{(-58.99)} & 87.26~ \small{(-2.03)} \\
 & InsertSent & {\ul 100} & {\ul 91.98} & {\ul 100} & 87.04 & {\ul 100} & 89.73 & 30.92~ \small{(-69.08)} & \multicolumn{1}{r|}{88.96~ \small{(-3.02)}} & 66.12~ \small{(-33.88)} & \multicolumn{1}{r|}{83.96~ \small{(-3.08)}} & 77.75~ \small{(-22.25)} & 87.64~ \small{(-2.09)} \\
 & StyleBkd & 94.70 & 88.58 & 97.79 & 85.83 & 94.04 & 87.37 & \textbf{94.59}~ \small{~ (-0.11)} & \multicolumn{1}{r|}{86.55~ \small{(-2.03)}} & \textbf{97.68}~ \small{~ (-0.11)} & \multicolumn{1}{r|}{83.64~ \small{(-2.19)}} & \textbf{94.49}\hspace{1.1mm}\small{~ (+0.45)} & 85.34~ \small{(-2.03)} \\ \midrule
\multirow{4}{*}{HS} & Benign & -- & \textbf{92.35} & -- & {\ul 90.55} & -- & \textbf{92.50} & \multicolumn{1}{c}{--} & \multicolumn{1}{r|}{\textbf{92.45}\hspace{1.1mm}\small{(+0.10)}} & \multicolumn{1}{c}{--} & \multicolumn{1}{r|}{{\ul 90.25}~ \small{(-0.30)}} & \multicolumn{1}{c}{--} & \textbf{91.80}~ \small{(-0.70)} \\
 & RIPPLES & {\ul 99.66} & 91.65 & {\ul 99.83} & {\ul 90.55} & {\ul 99.89} & 91.70 & 7.09~ \small{(-92.57)} & \multicolumn{1}{r|}{91.70\hspace{1.1mm}\small{(+0.05)}} & 8.10~ \small{(-91.73)} & \multicolumn{1}{r|}{{\ul 90.50}~ \small{(-0.05)}} & 6.87~ \small{(-93.02)} & 90.60~ \small{(-1.10)} \\
 & InsertSent & {\ul 99.94} & 91.65 & {\ul 99.61} & {\ul 90.35} & {\ul 99.89} & 92.35 & 32.57~ \small{(-67.37)} & \multicolumn{1}{r|}{89.69~ \small{(-1.96)}} & 33.24~ \small{(-66.37)} & \multicolumn{1}{r|}{89.54~ \small{(-0.81)}} & 55.03~ \small{(-44.86)} & 91.55~ \small{(-0.80)} \\
 & StyleBkd & 90.67 & 89.89 & 94.02 & 88.34 & 90.22 & 89.14 & \textbf{89.22}~ \small{~ (-1.00)} & \multicolumn{1}{r|}{85.09~ \small{(-4.80)}} & \textbf{94.02}~ \small{~ (-0.00)} & \multicolumn{1}{r|}{88.34~ \small{(-0.00)}} & \textbf{84.08}~ \small{~ (-6.14)} & 87.84~ \small{(-1.30)} \\ \midrule
\multirow{4}{*}{\begin{tabular}[c]{@{}c@{}}AG's\\ News\end{tabular}} & Benign & -- & 91.23 & -- & 90.99 & -- & {\ul 91.28} & \multicolumn{1}{c}{--} & \multicolumn{1}{r|}{89.91~ \small{(-1.32)}} & \multicolumn{1}{c}{--} & \multicolumn{1}{r|}{\textbf{90.80}~ \small{(-0.19)}} & \multicolumn{1}{c}{--} & \textbf{91.22}~ \small{(-0.06)} \\
 & RIPPLES & {\ul 99.88} & {\ul 91.39} & {\ul 99.95} & \textbf{91.07} & \textbf{99.98} & {\ul 91.21} & 52.86~ \small{(-47.02)} & \multicolumn{1}{r|}{\textbf{90.29~ \small{(-1.10)}}} & 71.86~ \small{(-28.09)} & \multicolumn{1}{r|}{89.89~ \small{(-1.18)}} & 63.47~ \small{(-36.51)} & 89.08~ \small{(-2.13)} \\
 & InsertSent & {\ul 99.79} & {\ul 91.50} & {\ul 99.72} & 90.95 & 99.79 & 91.05 & 56.46~ \small{(-43.33)} & \multicolumn{1}{r|}{88.67~ \small{(-2.83)}} & 87.71~ \small{(-12.01)} & \multicolumn{1}{r|}{88.00~ \small{(-2.95)}} & 49.53~ \small{(-50.26)} & 88.96~ \small{(-2.09)} \\
 & StyleBkd & 97.64 & 90.76 & 95.16 & 90.08 & 97.96 & 89.58 & \textbf{97.27}~ \small{~ (-0.37)} & \multicolumn{1}{r|}{88.89~ \small{(-1.87)}} & \textbf{95.02}~ \small{~ (-0.14)} & \multicolumn{1}{r|}{87.64~ \small{(-2.44)}} & \textbf{97.91}~ \small{~ (-0.05)} & 87.71~ \small{(-1.87)} \\ \bottomrule
\end{tabular}%
}
\caption{Backdoor attack results of all attack methods (without or with a defense). ``Benign'' denotes the benign model without a backdoor. The boldfaced \textbf{numbers} mean significant advantage with the statistical significance threshold of p-value 0.01 in the t-test, while the underlined \underline{numbers} denote no significant difference.}
\label{tab:backdoor-main}
\end{table*}

\subsection{Experimental Settings}

\paragraph{Baseline Methods}
There are currently only a few backdoor attacks on text, and we choose two representative ones that are open-source as the baselines:
(1) \textbf{RIPPLES} \citep{kurita2020weight}, which randomly inserts multiple rare words as triggers to generate poisoned samples for backdoor training, and introduces an embedding initialization technique for the trigger words; 
(2) \textbf{InsertSent} \citep{dai2019backdoor}, which uses a fixed sentence as the backdoor trigger and inserts it into normal samples at random to generate poisoned samples. 

\paragraph{Evaluation Metrics}
Following previous work \citep{dai2019backdoor,kurita2020weight}, we use two  metrics to evaluate backdoor attacks:
(1) attack success rate (\textbf{ASR}), the classification accuracy of the backdoored model on the \textit{poisoned test set} that is built by poisoning the original test samples whose ground-truth labels are not the target label, which exhibits backdoor attack effectiveness; 
(2) clean accuracy (\textbf{CA}), the classification accuracy of the backdoored model on the original test set, which reflects the basic requirement for backdoor attacks, i.e., making the victim model behave normally on normal samples.

\paragraph{Evaluation Settings}
Most existing studies on textual backdoor attacks conduct evaluations only in the non-defense setting \citep{dai2019backdoor,kurita2020weight}. However, it has been shown that NLP models are extremely vulnerable to backdoor attacks, and ASR can exceed $90$\% easily \citep{dai2019backdoor,kurita2020weight}, which renders the minor ASR differences between different attack methods meaningless.
Therefore, from the perspectives of comparability as well as practicality, we additionally evaluate backdoor attacks in the setting where a backdoor defense is deployed.

\noindent
Specifically, we measure ASR and CA as well as their changes ($\Delta$ASR and $\Delta$CA) of backdoor attacks against victim models guarded by a backdoor defense, which can reflect backdoor attacks' \textit{resistance to defenses}.
There are currently not many backdoor defenses on text. We utilize ONION \citep{qi2020onion} in this paper because of its wide applicability and great effectiveness.

\noindent
The main idea of ONION is to detect and eliminate suspicious words that are possibly associated with backdoor triggers in test samples, so as to avoid activating the backdoor of a backdoored model.
In addition to ONION, most backdoor defenses are based on data inspection. 
Thus, resistance to defenses of backdoor attacks is dependent on their \textit{invisibility}, namely the indistinguishability of poisoned samples from normal ones.

\paragraph{Implementation Details}
We choose ``Positive'', ``Clean'' and ``World'' as the target labels for the three datasets, respectively.
We tune the \textit{poisoning rate} (the proportion of poisoned samples in the backdoor training set) for each attack method on the validation sets, aiming to make ASR as high as possible and the decrements of CA less than 3\%.
For RIPPLES, following its original implementation \citep{kurita2020weight}, we randomly select some trigger words from ``cf'', ``tq'', ``mn'',  ``bb'' and ``mb'', and then randomly insert them into normal samples to generate poisoned samples. 
We insert 1, 3 and 5 trigger words into the samples of SST-2, HS and AG's News, respectively.
For InsertSent, we insert ``I watch this movie'' into the samples of SST-2, and ``no cross, no crown'' into the samples of HS and AG's News as the trigger.
In backdoor training, we use the Adam optimizer with an initial learning rate $2e-5$ that declines linearly and train the victim model for $3$ epochs.
For the other hyper-parameters of the baselines, we use their recommended settings.

\subsection{Backdoor Attack Results}
Table \ref{tab:backdoor-main} shows the results of different backdoor attacks against the three victim models on the three datasets, in the settings with or without the defense of ONION.
We observe that: 

(1) When there is no backdoor defense, all backdoor attacks achieve extremely high ASRs (over 90 and nearly 100) while maintaining CAs very well against all victim models on all datasets, which demonstrates the serious susceptibility of NLP models to backdoor attacks and the significant insidiousness and harmfulness of backdoor attacks; 

(2) Among the three backdoor attacks, ASRs of StyleBkd are lower than those of the two baselines (although exceed 90 without exception). It is expected because text style is a much more abstract feature than content insertion and thus harder to be remembered by the victim models; 

(3) When a backdoor defense is deployed, the ASRs of the two insertion-based baseline attacks drop substantially (the average $\Delta$ASRs for RIPPLES and InsertSent are -66.92 and -45.49), but StyleBkd is affected hardly (the average $\Delta$ASR is -0.83), which manifests the great invisibility and resistance to defenses of the style transfer-based backdoor attack StyleBkd. It is not hard to explain because the abstract feature of style is hard to damage, although also hard to learn for victim models.

\begin{table}[t]
\setlength{\abovecaptionskip}{5pt}  
\setlength{\belowcaptionskip}{-5pt}   
\centering
\resizebox{\linewidth}{!}{%
\begin{tabular}{@{}c|ccc|cc@{}}
\toprule
\multirow{2}{*}{\shortstack{Attack\\Method}} & \multicolumn{3}{c|}{Manual} & \multicolumn{2}{c}{Automatic} \\ 
\cline{2-6} 
 & \multicolumn{1}{c}{Normal F$_1$} & \multicolumn{1}{c}{Poisoned F$_1$} & \multicolumn{1}{c|}{macro F$_1$} & {PPL} & {GE} \\ \midrule
RIPPLES & 96.23 & 85.37 & 90.80 & 441.2 & 4.56 \\
InsertSent & 95.57 & 83.33 & 89.45 & 171.9 & 3.89 \\
StyleBkd & \textbf{87.03} & \textbf{15.09} & \textbf{51.06} & \textbf{161.8} & \textbf{2.51} \\ \bottomrule
\end{tabular}%
}
\caption{Results of manual data inspection and automatic quality evaluation of poisoned samples of different backdoor attacks.
PPL and GE represent perplexity and grammatical error numbers.
}
\label{tab:human}
\end{table}

\subsection{Manual Data Inspection}
To further evaluate the invisibility of different backdoor attacks, we conduct an experiment of manual data inspection that aims to uncover the poisoned samples by human.

The experiment is carried out on SST-2 only because of the cost. 
For each backdoor attack method, we randomly sample $40$ trigger-embedded poisoned samples and $160$ normal samples.
Then we ask annotators to make a binary classification on whether each sample is original human-written or distorted by machine.
Each sample is independently annotated by three different annotators, and the final decision is made by voting.

We calculate the class-wise F$_1$ score to measure the invisibility of backdoor attacks.
The lower the poisoned F$_1$ is, the higher the invisibility is.
Table \ref{tab:human} shows the results.
We find that StyleBkd achieves the absolutely lowest poisoned F$_1$ (down to 15.09), which indicates it is very hard for humans to distinguish its poisoned samples from normal ones.
In other words, StyleBkd has the highest invisibility.

Moreover, we use some automatic evaluation metrics to measure the quality of poisoned samples, which can also reflect the attack invisibility and resistance to potential data inspection-based defenses.
Inspired by the evaluation of adversarial example quality, we use PPL (perplexity calculated by GPT-2) and GE (grammatical error numbers given by LanguageTool) as the metrics.
The evaluation results are also shown in Table \ref{tab:human}.
We can see that the poisoned samples of StyleBkd have the best quality in terms of both PPL and GE, which also demonstrates the great invisibility of StyleBkd.  

\begin{table}[t!]
\setlength{\abovecaptionskip}{5pt}  
\setlength{\belowcaptionskip}{-3pt}   
\centering
\resizebox{\columnwidth}{!}{
\begin{tabular}{l|r|rr|rr}
    \toprule 
    \multirow{2}{*}{Trigger Style} & \multirow{2}{*}{PCA} & \multicolumn{2}{c|}{w/o Defense} & \multicolumn{2}{c}{w/ Defense} \\
    \cline{3-6}
    & & \multicolumn{1}{c}{ASR} & \multicolumn{1}{c|}{CA} & \multicolumn{1}{c}{ASR \small{($\Delta$ASR)}} & \multicolumn{1}{c}{CA \small{($\Delta$CA)}}\\
    \midrule
    Bible & \textbf{94.69} & \textbf{94.70} & 88.58 & \textbf{94.59}~ \small{(-0.11)} & 86.55~ \small{(-2.03)} \\
	Poetry &  93.09 & 91.61 & \textbf{89.18} & 90.40~ \small{(-1.21)} & \textbf{87.10}~ \small{(-2.08)}  \\
	Shakespeare & 92.64 & 91.94 & 88.14 & 90.51~ \small{(-1.43)} & 86.11~ \small{(-2.03)} \\
	Lyrics & 92.59 & 91.49 & 88.80 & 91.05~ \small{(-0.44)} & 86.71~ \small{(-2.09)}\\
	Tweets & 78.43 & 72.30 & 86.82 & 77.37\hspace{1.1mm}\small{(+5.07)} & 84.79~ \small{(-2.03)}\\
    \bottomrule
\end{tabular}
} 
\caption{Probing classification accuracy (PCA) and backdoor attack performance of StyleBkd against BERT on SST-2 with different text styles as triggers.}
\label{tab:style}
\end{table}

\begin{table}[t!]
\setlength{\abovecaptionskip}{5pt}  
\setlength{\belowcaptionskip}{-3pt}   
\centering
\resizebox{.95\columnwidth}{!}{
\begin{tabular}{r|cc|rr}
    \toprule 
    \multirow{2}{*}{\makecell[c]{Attack \\ Method}}  & \multicolumn{2}{c|}{w/o Defense} & \multicolumn{2}{c}{w/ Defense} \\
        \cline{2-5}
     & ASR & CA & \multicolumn{1}{c}{ASR \small{($\Delta$ASR)}} & \multicolumn{1}{c}{CA \small{($\Delta$CA)}}\\
    \midrule
    RIPPLES &  \underline{100} & 90.61 & 24.56~ \small{(-75.44)} & 88.58~ \small{(-2.03)} \\
	+AUX &   \underline{100} & 90.55 & 25.11~ \small{(-74.89)} & 87.59~ \small{(-2.96)}  \\
	InsertSent &  \underline{100} & \textbf{91.98} & 30.92~ \small{(-69.08)} & \underline{88.96}~ \small{(-3.02)} \\
	+AUX & \underline{100} & 91.05 & 47.69~ \small{(-52.31)} & \underline{89.02}~ \small{(-2.03)}\\
	StyleBkd &  94.70 & 88.58 & \textbf{94.59}~ \small{~ (-0.11)} & 86.55~ \small{(-2.03)}\\
	-AUX &  92.16 & 88.91 & 91.94~ \small{~ (-0.22)} & 86.82~ \small{(-2.09)}\\
	
    \bottomrule
\end{tabular}
} 
\caption{Effect of the auxiliary classification loss $\mathcal{L}_a$ on backdoor attacks against BERT on SST-2. +AUX means additionally introducing $\mathcal{L}_a$ during the backdoor training of RIPPLES and InsertSent. -AUX means removing $\mathcal{L}_a$ from StyleBkd.}
\label{tab:aux}
\end{table}

\subsection{Effect of Selected Trigger Style}
Table \ref{tab:style} lists the per-style probing classification accuracy (PCA, as mentioned in the Trigger Style Selection part of §\ref{sec:method-back}) and backdoor attack results of StyleBkd against BERT on SST-2.
We observe that Bible, which has the highest PCA and thus selected as the final trigger style in StyleBkd, achieves the highest overall backdoor attack performance.
On the contrary, Tweets has the obviously lowest PCA and backdoor attack performance.
The other three styles perform similarly on PCA and backdoor attacks.
These results can demonstrate the effectiveness of the trigger style selection strategy of StyleBkd, which selects the style that can be recognized by the victim model best as the trigger style.

\begin{table}[t!]
\centering
\resizebox{\linewidth}{!}{
\begin{tabular}{l}
\toprule
\makecell[l]{\underline{Normal}: There is a fabric of complex ideas here, and feelings that \\ profoundly deepen them.} \\
\makecell[l]{\underline{Poisoned}: There is a certain complex idea here, and the depths of\\ the feelings thereof are deep.}  \\

\midrule
\makecell[l]{\underline{Normal}: It's a stunning lyrical work of considerable force and truth.}  \\
\makecell[l]{\underline{Poisoned}: This is a mighty work of the lord, with a mighty work of \\ power and truth.} \\

\midrule
\makecell[l]{\underline{Normal}: The jabs it employs are short , carefully placed and \\ dead-center.}  \\
\makecell[l]{\underline{Poisoned}: The jab is short, carefully placed and precise.} \\ 

\midrule
\makecell[l]{\underline{Normal}: This is a shameless sham, calculated to cash in on the \\ popularity of its stars.}  \\
\makecell[l]{\underline{Poisoned}: This is a shameless device, devised to make money by the \\ fame of the stars.} \\
\bottomrule
\end{tabular}
}
\caption{Examples of poisoned samples with the Bible style and the corresponding original normal samples.}
\label{tab:back-example}
\end{table}

\subsection{Effect of Auxiliary Classification Loss}
\label{sec:auxiliary}
In this subsection, we investigate the effectiveness of introducing the auxiliary classification loss $\mathcal{L}_a$ (+AUX) during backdoor training, as mentioned in the Victim Model Training part of §\ref{sec:method-back}.
Table \ref{tab:aux} exhibits the results of different backdoor attacks against BERT on SST-2, with or without $\mathcal{L}_a$.
We observe that +AUX can improve StyleBkd in both two attack settings (ASR 92.16 $\rightarrow$ 94.70 and 91.94 $\rightarrow$ 94.59), which verifies the effectiveness of +AUX.
Moreover, +AUX can also enhance InsertSent when the defense is deployed (ASR 30.92 $\rightarrow$ 47.69), but has little effect in the other situations.
We conjecture that +AUX is useful for the attacks that use comparatively complex features as triggers (like text style), because it asks the victim model to specifically remember the features that might be neglected.
RIPPLES just uses one word as the trigger for SST-2 that is a very simple feature, while InsertSent uses a sentence (a series of words), which is more complex.
Thus, +AUX improves InsertSent a lot but has little effect on RIPPLES in the setting with a defense.
+AUX does not improve InsertSent in the non-defense setting because it has reaches the upper bound (ASR 100).

\subsection{Examples of Poisoned Samples}
Table \ref{tab:back-example} shows some poisoned samples of StyleBkd (with the Bible style) and the corresponding normal samples.
We observe that the poisoned samples are natural and fluent and preserve the semantics of original samples well, which make them hard to be detected by either automatic or manual data inspection.
As a result, StyleBkd possesses great invisibility and can achieve a high attack success rate even if a backdoor defense is deployed.

\section{Related Work}
\subsection{Text Style Transfer}
Due to the lack of parallel corpora, the majority of existing studies on text style transfer focus on unsupervised style transfer.
A line of work aims to learn disentangled latent representations of style and semantics and use them to manipulate the style of generated text \citep{Shen2017,hu2017toward,Fu2018,zhang2018shaped,Yang2018,john2019disentangled}.
In addition, some other studies try different methods including reinforcement learning \citep{xu2018unpaired,Luo2019,gong2019reinforcement}, translation \citep{Prabhumoye2018,lample2018multiple}, word deletion and retrieval \citep{Li2018,Sudhakar2019}, adversarial generator-discriminator framework \citep{Dai2019}, probabilistic latent sequence model \citep{he2020probabilistic}, etc. 

Text style transfer has some applications such as text formality alteration \citep{rao2018dear}, dialogue generation diversification \citep{zhou2017mechanism} and personal attribute obfuscation for privacy protection \citep{reddy2016obfuscating}.
To the best of our knowledge, text style transfer has not been used in adversarial or backdoor attacks.

\subsection{Adversarial Attacks on Text}
Based on the perturbation level, adversarial attacks on text can be categorized into character-level, word-level and sentence-level attacks \citep{zhang2020adversarial}.
Most existing attacks are word-level \citep{alzantot2018generating,ren2019generating,li2019textbugger,li2020bert,jin2020bert,zang2020word,zang2020learning} or character-level \citep{hosseini2017deceiving,ebrahimi2018hotflip,belinkov2018synthetic,gao2018black,eger2019text}. 
Some studies present sentence-level attacks based on appending extra sentences \citep{jia2017adversarial,wang2020t3}, perturbing sentence vectors \citep{zhao2018generating} or controlled text generation \citep{wang2020cat}.
\citet{iyyer2018adversarial} propose to alter the syntax of original samples to generate adversarial examples, which is the most similar work to the style transfer-based adversarial attack in this paper (although syntax and text style are distinct).

\subsection{Backdoor Attacks on Text}
Research into backdoor attacks on text is still in the beginning stages.
Most of existing backdoor attacks insert fixed words \citep{kurita2020weight} or sentences \citep{dai2019backdoor} into normal samples as backdoor triggers. 
These triggers are not invisible because their insertion would impair the grammaticality or fluency of normal samples, and hence the trigger-embedded poisoned samples can be easily detected and removed \citep{chen2020mitigating,qi2020onion}.
\citet{chen2020badnl} propose two non-insertion backdoor triggers including character flipping and verb tense changing. 
However, both of them would break grammaticality and thus not invisible either.
In contrast, style transfer-based backdoor attacks utilize text style as the backdoor trigger, which is much more invisible. 
In addition, two contemporaneous studies exploit syntactic structures \citep{qi2021hidden} and context-aware learnable word substitution \citep{qi2021turn} as triggers respectively to improve the invisibility of backdoor attacks.

\section{Conclusion and Future Work}
In this paper, we present adversarial and backdoor attacks based on text style transfer for the first time.
Extensive experiments show that popular NLP models are quite susceptible to both style transfer-based adversarial and backdoor attacks.
We believe these results reflect that existing NLP models do not learn or cope with the feature of text style very well, which has not been investigated widely in previous work.
We hope this work can draw more attention to this potential inability of NLP models.

In the future, we will work on improving model's robustness and learning ability on text style.
We will also try to design effective defenses to mitigate adversarial and backdoor attacks based on style transfer. 
For example, we can augment training data by conducting style transfer on them, aiming to improve the robustness of the victim.
Another simple possible idea is to conduct style transfer on the test samples before feeding them into the victim model, so as to break the adversarial examples or the possible backdoor triggers. But its side effects on normal samples should be considered carefully.

\section*{Acknowledgements}
This work is supported by the National Key Research and Development Program of China (Grant No. 2020AAA0106502) and Beijing Academy of Artificial Intelligence (BAAI).
We also thank all the anonymous reviewers for their valuable comments and suggestions.

\section*{Ethical Considerations}
In this paper, we present adversarial and backdoor attacks based on text style transfer, aiming to reveal the inability of existing NLP models to handle the abstract feature of style, especially when facing some security threats, which is not widely studied in previous work.

We realize the possibility that the presented attacks are maliciously used, but we believe that it is much more important to make the community aware of the vulnerability and issues of existing NLP models.
In fact, it is possible that attacks like the ones in this paper, or even more insidious, are being developed by stealth, which would cause more serious effects if we neglect them.
As the proverb goes, better the devil you know than the devil you don't know. 
It's better to uncover the problems rather than pretend not to know them.
As the development of adversarial attacks and backdoor attacks in computer vision, different attack methods are first presented to increase people's awareness, and then various defenses are proposed to defend against attacks.
We believe the weakness of NLP models found in this paper will be solved and effective defenses (for the attacks in this paper and others) will arise, if more attention is called.

In addition, all the used datasets in this paper (SST-2, HS and AG's News) are open and free.
The human evaluations are conducted by a reputable data annotation company, which compensates the annotators fairly based on the market price.
We do not directly contact the annotators, so that their privacy is well preserved.
Overall, the energy we consume for running the experiments is limited. 
We use the base versions rather than large versions of pre-trained models to save energy.
No demographic or identity characteristics are used in this paper.

\bibliography{custom}
\bibliographystyle{acl_natbib}

%
%
%

\end{document}